\documentclass[journal,transmag]{IEEEtran}

\usepackage{epsfig}
\usepackage{graphicx}

\usepackage[pagebackref=true,breaklinks=true,letterpaper=true,colorlinks,bookmarks=false]{hyperref}

\usepackage[linesnumbered,algo2e,boxed]{algorithm2e}
\graphicspath{{Figures/}}
\usepackage[figuresright]{rotating}
\usepackage{amsmath,amssymb,textcomp,subfigure,multirow,upgreek}
\usepackage{booktabs}
\usepackage{color,mdwlist}

\newcommand{\hao}[1]{\textcolor{black}{#1}}

\usepackage{enumitem}
\setitemize{noitemsep,topsep=0pt,parsep=0pt,partopsep=0pt}

\usepackage{caption,color}

\ifCLASSINFOpdf
\else
\fi

\begin{document}
%
\title{Facial Expression Translation using Landmark Guided GANs}


\author{\IEEEauthorblockN{Hao Tang and  
		Nicu Sebe 
	}
\thanks{
	Hao Tang is with the Department of Information Technology and Electrical Engineering, ETH Zurich,  Zurich 8092, Switzerland. E-mail: hao.tang@vision.ee.ethz.ch
	
    Nicu Sebe is with the Department of Information Engineering and Computer Science (DISI), University of Trento, Trento 38123, Italy. E-mail: sebe@disi.unitn.it.	
}
}

\markboth{IEEE Transactions on Affective Computing}%
{Shell \MakeLowercase{\textit{et al.}}: Bare Demo of IEEEtran.cls for IEEE Transactions on Magnetics Journals}
%

\IEEEtitleabstractindextext{%

\begin{abstract}
We propose a simple yet powerful Landmark guided Generative Adversarial Network (LandmarkGAN) for the facial expression-to-expression translation using a single image, which is an important and challenging task in computer vision since the expression-to-expression translation is a non-linear and non-aligned problem. Moreover, it requires a high-level semantic understanding between the input and output images since the objects in images can have arbitrary poses, sizes, locations, backgrounds, and self-occlusions. To tackle this problem, we propose utilizing facial landmark information explicitly. Since it is a challenging problem, we split it into two sub-tasks, (i) category-guided landmark generation, and (ii) landmark-guided expression-to-expression translation.  Two sub-tasks are trained in an end-to-end fashion that aims to enjoy the mutually improved benefits from the generated landmarks and expressions. Compared with current keypoint-guided approaches, the proposed LandmarkGAN only needs a single facial image to generate various expressions. Extensive experimental results on four public datasets demonstrate that the proposed LandmarkGAN achieves better results compared with state-of-the-art approaches only using a single image. The code is available at~\url{https://github.com/Ha0Tang/LandmarkGAN}.
\end{abstract}

}

\maketitle

\IEEEdisplaynontitleabstractindextext

%
\IEEEpeerreviewmaketitle

\section{Introduction}
In this paper, we mainly study how to enable machines to perform the facial expression-to-expression translation task, which is a classic task in  computer vision.
This task has many applications such as human-computer interactions, entertainment, and virtual reality.
Another important benefit of this task is that they can augment training data by generating images with given input images, which could be employed to improve the expression recognition accuracy. 
However, this task is difficult since (i) it needs to handle complex backgrounds with different illumination conditions, objects, and occlusions. (ii) it needs a high-level semantic understanding of the mapping between the input image and the output image since the objects in the inputs images can have arbitrary poses, sizes, locations, backgrounds, and self-occlusions.

Recently, Generative Adversarial Networks (GANs)~\cite{goodfellow2014generative} have shown the potential to solve this challenging task, and it can be utilized, for example, to convert a face with a neutral expression into different expressions.
GANs have demonstrated promising results in many generative tasks such as photo-realistic image generation \cite{karras2017progressive,wang2018high,park2019semantic,zhang2019self,brock2019large,shaham2019singan}.
Moreover, impressive image-to-image translation results have been obtained by using Conditional GAN (CGAN) \cite{isola2017image}, in which a conditional image is taken as input and the model outputs another image with a different style.
Given $m$ image domains, Pix2pix \cite{isola2017image} needs to train $m(m{-}1)$ models, which is inefficient and ineffective.
Recently, ComboGAN \cite{anoosheh2017combogan} and StarGAN \cite{choi2017stargan} are proposed to solve multi-domain image-to-image translation problem.
ComboGAN \cite{anoosheh2017combogan} requires $m$ models and StarGAN \cite{choi2017stargan} only needs to train one model. 
However, these methods cannot handle some specific image translation tasks such as pose generation \cite{ma2017pose,tang2019cycle} and gesture generation \cite{tang2018gesturegan} since person/gesture can have arbitrary poses, sizes, appearances, and locations in the wild, leading to infinity image domains. 

To address these limitations, several works have been proposed to generate images based on object keypoint or human skeleton.
For instance, Reed et al.~\cite{reed2016learning} proposed the Generative Adversarial What-Where Network (GAWWN), which generates birds conditioned on both text descriptions and object location.
Reed et al. \cite{reed2016generating} presented an extension of Pixel Convolutional Neural Networks (PixelCNN) to generate images part keypoints and text descriptions.
Ma et al. \cite{ma2017pose} proposed a two-stage reconstruction pipeline that generates novel person images.
Korshunova et al. \cite{korshunova2016fast} used facial keypoints to define the affine transformations of the alignment and realignment steps for face swap. 

Unlike these methods, we focus on facial expression-to-expression tasks.
Wei et al.~\cite{wei2017every} proposed a Conditional MultiMode Network (CMM-Net) for landmark-guided smile generation.
Di et al.~\cite{di2017gp} proposed the Gender Preserving Generative Adversarial Network (GPGAN) to synthesize faces based on facial landmarks.
Qiao et al.~\cite{qiao2018geometry} presented the Geometry-Contrastive Generative Adversarial Network (GCGAN) to generate facial expression conditioned on geometry information of facial landmarks, which explicitly employs the facial landmark to control the appearances and locations of the facial action units.
In the task, the facial landmark is especially useful because different expressions have different facial action units that have different shapes of mouths, lips, eyes, and face contours.
The expressions are mainly encoded by the landmarks. In other words, different expressions produce different landmark shapes w.r.t the eyes, eyebrows, and lips. 
Therefore, we explicitly employ landmarks as the guidance for expression translation which can embed the locations of the eyes, eyebrows, and lips into the conditional image.

\begin{figure*}[!t]
	\centering
	\includegraphics[width=1\linewidth]{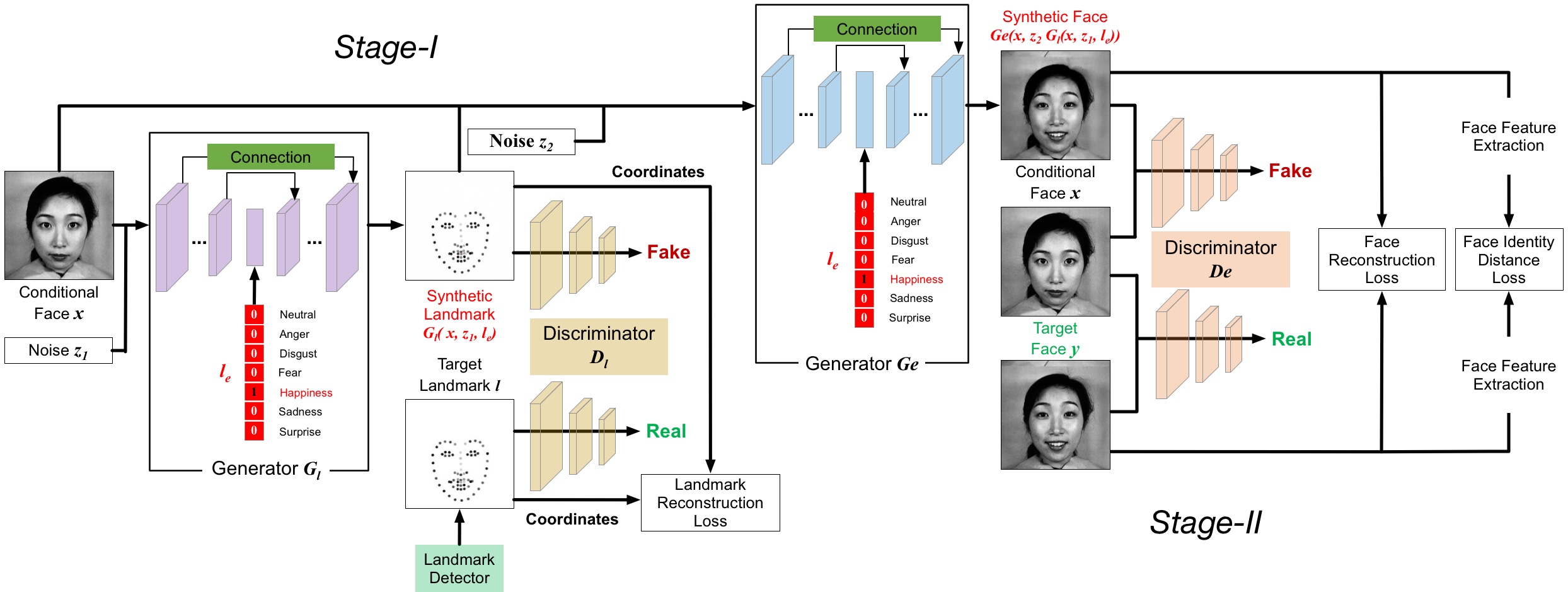}
	\caption{The pipeline of the proposed LandmarkGAN. The input of stage-I is a face image and the expression labels, and the output is the landmark image of the target face. The inputs of stage-II are a conditional image and the synthetic landmark image generated in the stage-I, and the output is a face with the target expression but preserves identity information.  The generators $G_l$ and $G_e$ try to generate landmark/image to fool the discriminator $D_l$ and $D_e$, respectively. The discriminators $D_l$ and $D_e$ learn to classify fake $G_l(x, z_1, l_e)$ and real landmark $l$, and fake $(G_e(x, z_2, G_l(x, z_1, l_e), l_e), x)$ and real image pair $(y, x)$, respectively.}
	\label{fig:framework}
\end{figure*}

But during the testing period the aforementioned methods employ a keypoint detector to extract the landmarks of the target object, which introduces an extra module and then reduces the flexibility of the proposed system.
To overcome these difficulties, we propose a novel LandmarkGAN in this paper which can generate landmarks and images without any keypoint detectors during the testing stage, which thus can be used in practical applications with less network overhead.

Specifically, we propose a novel LandmarkGAN framework, as shown in Figure~\ref{fig:framework}.
LandmarkGAN comprises two generation stages, (i) category-guided landmark generation and (ii) landmark-guided expression-to-expression translation.
In the first stage, we employ a CGAN model to generate the targeted object landmark conditioned on the object label and the input image. 
We can also generate the targeted object landmark conditioned on the object label and the landmark of the input image. 
However, this requires an additional object landmark detector to detect the landmarks of the input image.
Therefore, instead of using the detected landmarks of the input image, we use the input image directly as the conditioning prior.
To further generate more accurate landmarks, we use the Mean Squared Error (MSE) loss to calculate the errors on the coordinates of the landmarks.

In the second stage, the generated object landmarks and the conditional image are concatenated together as the input to a U-Net~\cite{isola2017image} to generate the targeted image with the corresponding facial expression.
Note that in the first stage, facial expressions are represented by one-hot vectors in which there is only one element set to 1 to represent the corresponding expression while the others are set to 0.
First, the one-hot vector is encoded by the fully connected layers and then the encoded label representations are concatenated with the image representations to generate the target facial landmarks.
Two stages are trained jointly by an end-to-end fashion, in this way, the generated landmark can help generate more photo-realistic faces, and high-quality faces then improve the generation of landmarks.

Compared with state-of-the-art keypoint-guided methods such as C2GAN \cite{tang2019cycle} which need image and keypoint as input during the testing stage, the proposed LandmarkGAN only needs a single face image to generate diverse landmarks and faces.
Extensive experiments on several public datasets demonstrate that LandmarkGAN can generate better results compared with state-of-the-art methods. 
Thus, these generated images by the proposed method can augment the training data and improve the performance of the facial expression classifier.

Overall, the contributions of this paper are summarized as follows:
\begin{itemize}
	\item We propose a novel LandmarkGAN model for facial expression-to-expression translation tasks. LandmarkGAN is comprised of two sub-modules, one for category-guided landmark generation and the other one for landmark-guided expression generation. Compared with current methods that need images and landmarks as input during the testing stage, the proposed LandmarkGAN only needs a single image to generate faces with different novel expressions.
	\item The proposed LandmarkGAN is trained in an end-to-end fashion that aims to enjoy the mutually improved benefits from the generated landmarks and expressions. To the best of our knowledge, we are the first to be able to make one GAN framework work on both facial expression image generation and facial landmark generation tasks.
	\item Experimental results on four challenging datasets demonstrate the superiority of LandmarkGAN over the state-of-the-art models on the unconstrained facial expression-to-expression task. Moreover, the generated images have high-quality and preserve the identity information, and thus they can be used to boost the performance of facial expression classifiers. 
\end{itemize}

\section{Related Work}
\label{sec:relatewprk}
\noindent\textbf{Generative Adversarial Networks (GANs)} \cite{goodfellow2014generative} have shown the capability of generating high-quality images \cite{karras2017progressive,wang2018high,park2019semantic,zhang2019self,brock2019large,shaham2019singan,zhang2017compositional,tan2021controllable,yu2020toward}.
GANs have two important components, i.e., generator $G$ and discriminator $D$.
The goal of $G$ is to generate photo-realistic images from a noise vector, while $D$ trying to distinguish between the real image and the image generated by $G$.
Although it is successful in many tasks, it also has challenges such as how to generate images of high visual fidelity and how to control the position and the shape of generated objects.

To generate some specific kind of images given the input for users' goal, Conditional GAN (CGAN) \cite{mirza2014conditional} is proposed.
A CGAN always combines a basic GAN and an external information, such as labels \cite{d2021ganmut,tang2019expression,tang2019attribute}, text descriptions \cite{reed2016learning,xu2022predict,tao2022df}, segmentation maps \cite{wu2022cross,wu2022crosspr,tang2022local,ren2021cascaded,tang2021layout,liu2021cross,tang2020dual,tang2020local,liu2020exocentric,tang2019multi}
and images \cite{isola2017image,zhu2017unpaired}.
For example, GANmut \cite{d2021ganmut} introduces a novel GAN-based framework that learns an expressive and interpretable conditional space to generate a gamut of emotions, using only the categorical emotion labels.

\noindent\textbf{Image-to-Image Translation} frameworks use input-output data to learn a parametric mapping between inputs and outputs, e.g., Isola et al.~\cite{isola2017image} proposed Pix2pix, which uses a CGAN to learn a translation function from input to output image domains.
Zhu et al.~\cite{zhu2017unpaired} introduced CycleGAN, which achieves unpaired image-to-image translation using the cycle-consistency loss.
However, existing image-to-image translation models (e.g., CycleGAN~\cite{zhu2017unpaired}, DiscoGAN \cite{kim2017learning}, DualGAN~\cite{yi2017dualgan}, Pix2pix \cite{isola2017image}, ComboGAN \cite{anoosheh2017combogan}, StarGAN \cite{choi2017stargan}, AttentionGAN \cite{tang2021attentiongan}) are inefficient and ineffective.
These approaches cannot handle some specific image generation tasks such as pose \cite{ma2017pose} and gesture \cite{tang2018gesturegan} generation, which could have infinity image domains since person/gesture can have arbitrary poses, sizes, appearances and locations in the wild.

\noindent\textbf{Keypoint-Guided Image-to-Image Translation.}
To address these limitations, several works have been proposed to generate images based on object keypoint~\cite{ge2018fd,dong2018soft,zhou2020makelttalk,chen2019hierarchical,zakharov2019few,dall2022graph,tang2021total,tang2020unified,tang2020bipartite,tang2020xinggan}.
For instance, Di et al.~\cite{di2017gp} proposed GPGAN to synthesize faces based on facial landmarks.
Reed et al.~\cite{reed2016learning} proposed GAWWN, which generates birds conditioned on both text descriptions and object location.
Ma et al. proposed PG2~\cite{ma2017pose}, which achieves person image translation using a conditional image and a target pose image.
Sun et al. \cite{sun2017natural} proposed a two-stage framework to  perform head inpainting conditioned on the generated facial landmark in the first stage.
Korshunova et al. \cite{korshunova2016fast} used facial keypoints to define the affine transformations of the alignment and realignment steps for face swap. 
Wei et al. \cite{wei2017every} proposed a Conditional MultiMode Network (CMM-Net) for landmark-guided smile generation.
Qiao et al.~\cite{qiao2018geometry} presented GCGAN to generate facial expression conditioned on geometry information of facial landmarks. 
Song et al. \cite{song2017geometry} proposed G2GAN for facial expression synthesis guided by fiducial points.
These methods employ object keypoints to guide the image generation process since the object keypoints  provide four types of information for generation at the same time, i.e., category, location, scale, and orientation of objects.

However, existing approaches such as \cite{qiao2018geometry,ma2017pose,tang2019cycle,song2017geometry} employ a keypoint detector to generate keypoints during the testing stage, which reduces the flexibility of the proposed system or method.
In this paper, we propose the LandmarkGAN which can generate facial landmarks and facial expressions without any keypoint detectors, which thus can be used in practical applications with less network overhead. Moreover,  our method can generate both facial landmarks and images (see Figure~\ref{fig:landmark}), which is not investigated in both existing methods \cite{qiao2018geometry,ma2017pose,tang2019cycle,song2017geometry}.

\section{LandmarkGAN  Formulation}
\label{sec:method}
The goal of GANs is to learn a mapping from random noise~$z$ to output image $y$, $G{:} z {\mapsto} y$~\cite{goodfellow2014generative}.
The generator $G$ is trained to generate images that ``fools'' the discriminator $D$.
The discriminator $D$, which is optimized to distinguish real from fake images.
The objective of the original GAN can be formalized as follows:
\begin{equation}
	\mathcal{L}_{GAN}(G, D) {=} 
	\mathbb{E}_y \left[ \log D(y) \right] {+} \mathbb{E}_{z} \left[\log (1 {-} D(G(z))) \right].
\end{equation}
While, for conditional GANs, which learn the mapping with a conditional image $x$, i.e., $G{:} [x, z] {\mapsto} y$.
To utilize $x$ during training phase, the objection can be updated to conditional version:
\begin{equation}
	\begin{aligned}
		\mathcal{L}_{cGAN}(G, D) & {=} \mathbb{E}_{x, y} \left[ \log D(x, y) \right] \\
		& {+}  \mathbb{E}_{x, z} \left[\log (1 {-} D(x, G(x, z))) \right],
	\end{aligned}
	\label{eqn:conditonalgan}
\end{equation}
where $D$ tries to maximize this objective while $G$ tries to minimize it.
Thus, the solution is $G^*{=}\arg \min \limits_G \max \limits_D \mathcal{L}_{cGAN}(G, D)$.
For our LandmarkGAN, which comprises two stags, i.e., facial landmark and facial expression generation (see Figure~\ref{fig:framework}).

\subsection{Stage I: Facial Landmark Generation}
During the first stage, the translation mapping can be formulated as $G_l{:} [x, z_1, l_e] {\mapsto} l$, where $x$ denotes the input image, $z_1$ represents a random noise, $l_e$ denotes expression label, and $l{=}\sum_{i=1}^{68}(p_i,q_i)$ denotes the target landmark.
For generating training data, we employ OpenFace \cite{Baltrusaitis2016} as facial landmark detector to extract 68 facial landmarks, and then encode these points as a color image.
Specifically, each pixel in a radius of $S_l{=}4$ around the corresponding facial landmark is filled with the color in the point of the corresponding image $x$ and 255 (white color) elsewhere.
This embedding strategy ensures that the model to learn both locations and colors information of facial landmarks. 
In this way, the facial landmarks not only provide the clues regarding facial shapes to the generator but also provide the color information of different facial parts to the generator.
Then, the objective of first stage can be formulated as follows,
\begin{equation} 
\begin{aligned}
\mathcal{L}_{cGAN}(G_l, D_l) & {=}  \mathbb{E}_{l} \left[\log D_l(l) \right] \\
& {+}\mathbb{E}_{x, z_1} \left[\log (1 {-} D_l(G_l(x, z_1, l_e))) \right],
\end{aligned}
\label{eqn:stage_1}
\end{equation}
where $l_e$ is the one-hot vector of expressions and $l_e$ is connected with the image embedding at the bottleneck fully connected layer.

In addition, some prior works have verified it favorable to combine the GAN loss with a more traditional reconstruction loss, such as $L_1$ loss \cite{isola2017image} and $L_2$ loss \cite{pathak2016context}.
Both  $L_1$ loss \cite{isola2017image} and $L_2$ loss \cite{pathak2016context} are calculated based on images,  while for improving the accuracy of the generated landmarks, we propose the landmark reconstruction loss. Unlike the general MSE loss that calculates on the pixels of the image, the  landmark reconstruction loss calculates on the coordinates of the landmarks, which can be expressed as,
\begin{equation}
\mathcal{L}_{Landmark}(G_l) {=} \mathbb{E}_{l, \hat{l}} \left[ \frac{1}{68}\sum_{i=1}^{68}\sqrt{(p_i-\hat{p_i})^2+(q_i-\hat{q_i})^2} \right],
\end{equation}
where $\hat{l}{=}G_l(x, z_1, l_e){=}\sum_{i=1}^{68}(\hat{p_i},\hat{q_i})$, $p$ and $q$ represents the $x$ and $y$ coordinate of the lamdmark $i$, respectively. 

Thus, the final loss of the first stage is,
\begin{equation}
\begin{aligned}
\mathcal{L}_{1} = \mathcal{L}_{cGAN}(G_l, D_l) + \lambda_1 \mathcal{L}_{Landmark}(G_l),
\end{aligned}
\label{eqn:overallloss_1}
\end{equation}
where the parameter $\lambda_1$ controls the relative importance of both $\mathcal{L}_{cGAN}(G_l, D_l)$ and $\mathcal{L}_{Landmark}(G_l)$.

\subsection{Stage II: Facial Expression Generation}
Facial expression generation is conditioned on the landmark image $G_l(x, z_1, l_e)$ generated in the first stage, a conditional image $x$, and expression label $l_e$.
We first define $[x, G_l(x, z_1, l_e), l_e]{=}\widehat{x}$,
thus the adversarial objective of the second stage can be expressed as,
\begin{equation}
\begin{aligned}
 \mathcal{L}_{LGAN}(G_e, D_e) & = \mathbb{E}_{x, y} \left[\log D_e(x, y) \right] \\
& +  \mathbb{E}_{\widehat{x}, x, z_2} [\log (1 - D_e(x, G_e(\widehat{x}, z_2))],
\end{aligned}
\label{eqn:clandmarkgan}
\end{equation}
in this way, the input image and the output image can be better aligned. Our model only needs to pay attention to the areas related to the expression change, and not to the difference unrelated to the expression change, thereby reducing the difficulty of learning.

Next, the face reconstruction loss of generator $G_e$  measured by $L1$ and $L2$ distances can be defined as
$\mathcal{L}_{L1}(G_e) {=} \mathbb{E}_{\widehat{x}, y, z_2} \left[ \left|\left| y - G_e(\widehat{x}, z_2) \right|\right|_1 \right]$
and 
$\mathcal{L}_{L2}(G_e) {=} \mathbb{E}_{\widehat{x}, y, z_2} \left[ \left|\left| y - G_e(\widehat{x}, z_2) \right|\right|_2 \right]$, respectively.
 
 \begin{figure}[!t]
 	\centering
 	\includegraphics[width=0.7\linewidth]{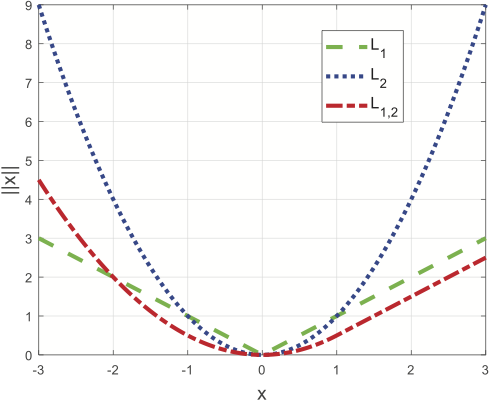}
 	\caption{Comparison of $L_1$, $L_2$ and $L_{1,2}$ losses on the image translation task.}
 	\label{fig:norm}
 \end{figure}

In this paper, we also explore an alternative option and propose a robust version loss in Figure~\ref{fig:norm}, i.e., $\mathcal{L}_{L{1,2}}$, which takes both $L1$ and $L2$ losses into consideration simultaneously.
The proposed $\mathcal{L}_{L{1,2}}$ loss is less sensitive to outliers than the $L2$ loss, and in some cases it could prevent exploding gradients \cite{girshick2015fast}.
Moreover, we observe that the $\mathcal{L}_{L{1,2}}$ loss can provide more image details than the $L1$ loss in the preliminary experiments.
Thus, the definition of the proposed $\mathcal{L}_{L{1,2}}$ is:

\begin{equation}
\begin{split}
& \mathcal{L}_{L{1,2}}(G_e)=
\begin{cases}
	0.5*\mathcal{L}_{L_2}(G_e) & if \left| y - G_e(\widehat{x}, z_2) \right| {<}  1, \\
	\mathcal{L}_{L_1}(G_e) {-} 0.5 & if \left| y - G_e(\widehat{x}, z_2) \right| {\geq}  1.
\end{cases}
\end{split}
\end{equation}
Note that the proposed $L_{1,2}$ loss is similar to the Huber loss, but we are the first to use this loss in the generation task.

Moreover, to preserve the face identity while expression synthesis, we propose the face identity distance loss, which can be expressed as follows,
\begin{equation}
\begin{aligned}
\mathcal{L}_{identity}(G_e) = \mathbb{E}_{\widehat{x}, y, z_2} \left[ \left|\left| F(y) - F(G_e(\widehat{x}, z_2)) \right|\right|_1 \right],
\end{aligned}
\label{eqn:preserve}
\end{equation}
where $F$ is a face feature extractor. 
We employ \cite{amos2016openface} to compare the identity distance between the target face $y$ and the generated face $G_e(\widehat{x}, z_2)$, and try to minimize the distance $F(y) {-} F(G_e(\widehat{x}, z_2))$ using $L_1$ distance. 
The model proposed in \cite{amos2016openface} is pre-trained as a classifier to distinguish between tens of thousands of identities, so it has the ability to capture the most prominent feature for face identity discrimination.

Therefore, our overall objective of the second stage is:
\begin{equation}
\begin{aligned}
\mathcal{L}_2 = \mathcal{L}_{LGAN}(G_e, D_e) + \lambda_2 \mathcal{L}_{L_{1,2}}(G_e) + \lambda_3\mathcal{L}_{identity}(G_e),
\end{aligned}
\label{eqn:overallloss}
\end{equation}
where the parameter $\lambda_2$ and $\lambda_3$ control the relative importance of $\mathcal{L}_{LGAN}(G_e, D_e)$, $\mathcal{L}_{L_{1,2}}(G_e)$ and $\mathcal{L}_{identity}(G_e)$. 
Thus, the full objective  of the proposed LandmarkGAN is $\mathcal{L} {=} \mathcal{L}_1 {+} \mathcal{L}_2$.
During the training stage, the full objective is optimized by an end-to-end fashion that aims to enjoy the mutually improved benefits from both stages.

In addition, in the first stage, the discriminator $D_l$ tries to distinguish landmark 
$l$ from $G_l(x, z_1, l_e)$.
In the second stage, the discriminator $D_e$ learns to classify fake $(G_e(\widehat{x}, z_2), x)$ and real $(y, x)$ pairs, as shown in Figure~\ref{fig:framework}.
Thus, the loss functions for the discriminator $D_l$ and $D_e$ are:
$\mathcal{L}(D_l) {=} \frac{1}{2}\left[ \mathcal{L}_{bce}(D_l(l), 1) {+} \mathcal{L}_{bce}(D_l(G_l(x_1,z_1,l_e), 0)\right],$ and 
$\mathcal{L}(D_e) {=} \frac{1}{2} [ \mathcal{L}_{bce}(D_e(y, x), 1) {+}  \mathcal{L}_{bce}(D_e(G_e(\widehat{x}, z_2), x), 0) ],$ respectively, 
where $\mathcal{L}_{bce}$ represents the Binary Cross Entropy loss function between the target and the output. 
We also divide the objective of discriminators by 2 to slow down the rate of discriminators $D_l$ and $D_e$ relative to generators $G_l$ and $G_e$. 
Moreover, Isola et al.~\cite{isola2017image} have found that the noise $z_1$ and~$z_2$ are not necessary in Pix2pix, thus in our LandmarkGAN, we follow Pix2pix~\cite{isola2017image} and provide noise only through dropout operation in generators.

\subsection{Implementation Details}
\noindent \textbf{LandmarkGAN Architecture.}
We employ U-Net \cite{isola2017image} for generators $G_l$ and $G_e$, U-Net actually is a network with skip connections between encoder and decoder, as shown in Figure~\ref{fig:framework}. 
Batch normalization \cite{ioffe2015batch} is employed except in the first layer of the encoders.
Leaky ReLU is used in the encoders with a slope of 0.2.
While all ReLUs in the decoders are not leaky.
And the last layer of the decoders is the Tanh activation function.
Moreover, the first three blocks of the decoders use the Dropout layer with a dropout rate of 50\%.
For discriminators $D_l$ and $D_e$, we adopt PatchGAN proposed in~\cite{isola2017image}, PatchGAN tries to classify if each $70 {\times} 70$ patch in an image is real or fake.
The final layer of discriminators employs a Sigmoid activation function.

\begin{figure*}[!t]
	\centering
	\setcounter{subfigure}{0}
	\subfigure{\includegraphics[width=.32\linewidth]{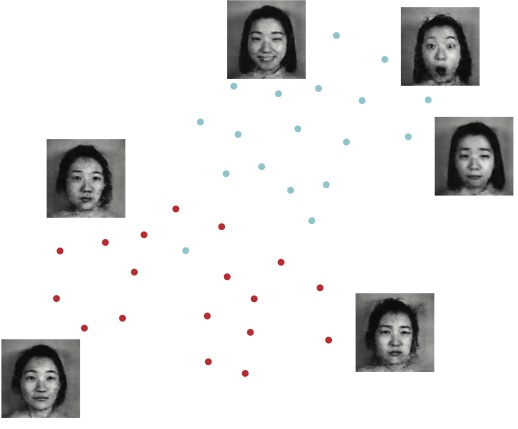}}
	\subfigure{\includegraphics[width=.32\linewidth]{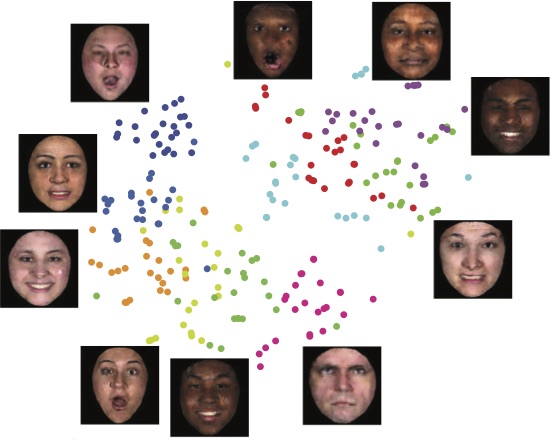}}
	\subfigure{\includegraphics[width=.32\linewidth]{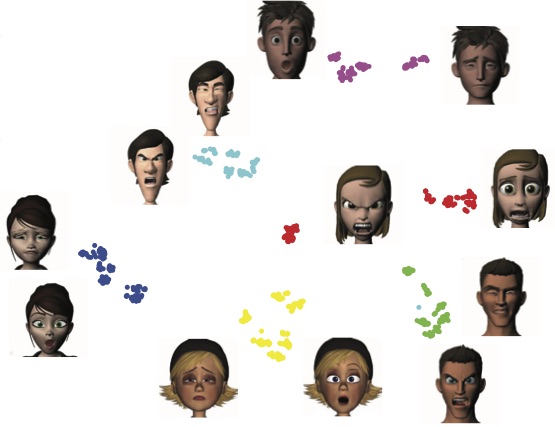}}
	\caption{Identity feature space of JAFFE (\textit{left}), 3DFE (\textit{middle}), and FERG (\textit{right}). Each color represents a subject.}
	\label{fig:plot_id}
\end{figure*}

\begin{table*}[!t]
	\centering
	\caption{Quantitative results of different components on JAFFE, 3DFE, and FERG. }
	\begin{tabular}{lccccccccc} \toprule
		\multirow{2}{*}{Setting}        &\multicolumn{3}{c}{PSNR $\uparrow$}      & \multicolumn{3}{c}{SSIM $\uparrow$}   & \multicolumn{3}{c}{IS $\uparrow$}  \\ \cmidrule(lr){2-4} \cmidrule(lr){5-7} \cmidrule(lr){8-10}
		&JAFFE       & 3DEF   & FERG   &JAFFE   &3DFE   &FERG       & JAFFE    & 3DFE    & FERG      \\ \midrule
		LandmarkGAN - D - I - LR     &16.8467    &22.3147&32.9560 &0.4968&0.8737 &0.9813  &1.4508 & 1.7392 &1.5576  \\
		LandmarkGAN - D - I             &17.2185    &22.8473&36.4789 &0.5197&0.8903 &0.9884  &1.4619   & 1.7565 &1.5756   \\  
		LandmarkGAN - LR                &17.1502     &22.8186&34.1283 &0.5186&0.8881 &0.9863  &1.4522   & 1.7489 &1.5846  \\   
		LandmarkGAN - I                   &17.2613    &23.1719&37.7746 &0.5244&0.9021 &0.9890   &1.4858    & 1.7610 &1.5937   \\  
		LandmarkGAN - D                  &17.2852    &23.8509&38.8294 &0.5361&0.9133 & 0.9930 & 1.5508 & 1.7625 &1.5960  \\  
		LandmarkGAN   ($L_{1,2}$, $l_e {\rightarrow} G_l \, \& \, l_e {\rightarrow} G_e$, w/ soft-crops)                    &\textbf{17.2911}&\textbf{23.8754}& \textbf{38.9132} &\textbf{0.5392}& \textbf{0.9332} & \textbf{0.9932} & \textbf{1.5708}  & \textbf{1.7984} &\textbf{1.6103} \\  \hline
		LandmarkGAN  ($L_{1}$, $l_e {\rightarrow} G_l  \, \& \, l_e {\rightarrow} G_e$, w/ soft-crops)    & 17.1892  & 23.4216 & 38.3760  & 0.5189 & 0.9174 & 0.9789 &  1.5687 & 1.7509 & 1.5762 \\
	    LandmarkGAN  ($L_{2}$, $l_e {\rightarrow} G_l  \, \& \, l_e {\rightarrow} G_e$, w/ soft-crops)    & 17.2343 & 23.6743 &  38.7598 & 0.5276 & 0.9265 & 0.9862 &  1.5268 & 1.7653 & 1.5876 \\	\hline
	    {LandmarkGAN   ($L_{1,2}$, $l_e {\rightarrow} G_l$, w/ soft-crops)} & {17.2807} & {23.8532}& {38.8750}& {0.5375}& {0.9324}& {0.9919}& {1.5698}& {1.7894}& {1.6062} \\
	    {LandmarkGAN ($L_{1,2}$, $l_e {\rightarrow} G_e$, w/ soft-crops)} & {17.0156} & {23.3642}& {37.5791}& {0.5108}& {0.9185}& {0.9868}& {1.5202}& {1.7530}& {1.5693}  \\ \hline
        LandmarkGAN   ($L_{1,2}$, $l_e {\rightarrow} G_l \, \& \, l_e {\rightarrow} G_e$, w/o soft-crops) & 17.0726 & {23.6752}& {38.6709}& {0.5167}& {0.9298}& {0.9912}& {1.5649}& {1.7785}&{1.6096}   \\
		\bottomrule	
	\end{tabular}
	\label{tab:ab}
\end{table*}  

\noindent \textbf{LandmarkGAN Optimization.}
We follow the same optimization method in \cite{goodfellow2014generative} to optimize the proposed LandmarkGAN, i.e., one gradient descent step on discriminator and generator alternately.
We first train $G_l$ and $D_l$ with $G_e$ and $D_e$  fixed, and then train $G_e$ and $D_e$ with $G_l$ and $D_l$ fixed.
The proposed LandmarkGAN is trained by an end-to-end fashion.
During the training stage, we employ OpenFace \cite{Baltrusaitis2016} to extract facial landmarks as training data.
\hao{We noticed that as the landmark detection accuracy decreases, the performance of the proposed LandmarkGAN decreases very slowly. This is because our model is also able to generate the correct landmark under the guidance of the expression label in the first stage. At the same time, in the second stage, our model can also generate the target image under the guidance of expression label. The most important thing is that we have the ground truth image, so it can correctly guide the model to learn by using the proposed face reconstruction loss, face identity distance loss, and adversarial loss.}

While during the testing stage, we use the trained generator $G_l$ as a facial landmark detector to generate the target landmark, which means LandmarkGAN does not need extra modules, while PG2~\cite{ma2017pose} and C2GAN \cite{tang2019cycle} need to employ extra module to produce landmark in order to generate variety expressions.
All images are scaled to $256{\times} 256$.
Each model is trained 200 epochs with batch size 1.
We employ the Adam \cite{kingma2014adam} with momentum terms  $\beta_1{=}0.5$ and $\beta_2{=}0.999$ as our solver.
The initial learning rate for Adam is 0.0002.

We set $S_l$=4, $\lambda_1$=2, $\lambda_2$=100 and $\lambda_3$=0.1 according to our ablation experiments.
For better training  the landmark generator, we adopt the pre-trained PDM decoder used in \cite{sun2017natural}.
We also employ dual discriminators as in 
Nguyen et al.~\cite{nguyen2017dual} and GestureGAN \cite{tang2018gesturegan}, which have demonstrated that they improve the ability of discriminator to generate more photo-realistic images.
The proposed LandmarkGAN is implemented by deep learning framework PyTorch \cite{paszke2019pytorch}.

\section{Experiments}
\label{sec:experiment}

\subsection{Experimental Setups}
\noindent \textbf{Datasets.}
To validate the effectiveness of the proposed LandmarkGAN, we conduct experiments of unconstrained facial expression-to-expression translation on four public datasets (i.e., JAFFE~\cite{lyons1998coding}, 3DFE~\cite{yin20063d}, FERG~\cite{aneja2016modeling} and RaFD~\cite{langner2010presentation}) for four different purposes.

(1) We evaluate on the Japanese Female Facial Expression (JAFFE) dataset~\cite{lyons1998coding} to see if our model still works with limited training samples, which will inspire few-shot image generation tasks. The JAFFE~dataset~\cite{lyons1998coding} contains 217 images of 6 basic facial expressions and 1 neutral posed by 10 Japanese females, and each subject has 3 images except one has 4 images.
We randomly select 2 subjects as testing data and the resting 8 people as training data. 

(2) The evaluation on the 3D Facial Expression (3DFE) dataset proves that our method has a good ability to control the intensity of expression.
The 3DFE dataset \cite{yin20063d} includes 100 subjects with 7 expressions and each expression has 4 levels of intensity, thus this dataset has 100 subjects $\times$ 7 expressions $\times$ 4 images = 2,800 images in total.
For each expression, we randomly choose 10 subjects as testing data and the resting 360 images as training samples.

(3) Our evaluation on the Facial Expression Research Group (FERG) dataset shows that our method is also effective on cartoon images.
The FERG dataset \cite{aneja2016modeling} is a dataset of stylized characters with 7 basic facial expressions, which contains about 50,000 images modeled using the MAYA software.
In our experiments, we randomly select 1,200 faces from each expression, and then randomly choose 100 images out of 1,200 images as testing data and the rest of the 1,100 images as training samples.

(4) Our evaluation on the Radboud Faces (RaFD) dataset proves that our method is more effective than the existing methods such as GANimation~\cite{pumarola2018ganimation} and C2GAN \cite{tang2019cycle} since both use RaFD as a standard dataset.
The RaFD dataset \cite{langner2010presentation}, which over 8,000 color face images collected from 67 subjects with eight different emotional expressions, i.e., anger, fear, disgust, sadness, happiness, surprise, neutral and contempt.
Each emotion has 1,005 images are captured from five cameras with different angles and each subject is asked to show three different gaze directions. 
For each emotion, we select 67\% images as training data and the rest 33\% images as testing data.
Similar to C2GAN~\cite{tang2019cycle}, all the images are rescaled to $256 {\times} 256 {\times} 3$ without any preprocessing in all experiments since we aim to conduct unconstrained facial expression-to-expression translation task.

These datasets contain faces with different races and styles, and they have different illuminations, occlusions, pose conditions, and backgrounds.

\noindent \textbf{Evaluation Metrics.} 
We employ Peak Signal-to-Noise Ratio (PSNR), Structural Similarity Index (SSIM)~\cite{wang2004image}, Inception Score (IS) \cite{salimans2016improved}, and LPIPS \cite{zhang2018perceptual} to measure the quality of generated images.
More specifically, PSNR measures the difference of two images from pixel level, while SSIM evaluates the similarity of two images from global level.
Both IS and LPIPS estimate generated images from a higher semantic level.
For all metrics except LPIPS, higher is better. 

\subsection{Experimental Results}
\noindent \textbf{Ablation Study.}
We first conduct an ablation study of the proposed LandmarkGAN and show the results without the Double discriminators strategy (D), face Identity distance loss (I), and Landmark Reconstruction loss (LR), respectively.
We use the scheme of training a dual-discriminator instead of one discriminator as a more stable way to improve the capacity of discriminators similar to~\cite{durugkar2016generative}. 
To be more specific, the dual-discriminator architecture can better approximate the optimal discriminator. 
If one of the discriminators is trained to be far superior over the generators, the generators can still receive instructive gradients from the other one.
Table \ref{tab:ab} shows the results using different components of LandmarkGAN on the JAFFE, 3DFE, and FERG datasets.
We observe that removing any terms from I, FR, LR, and D can degrade the generation performance.
Thus, we conclude that these items are important for generating photo-realistic images.

Moreover, we compare the proposed $L_{1,2}$ with $L_1$ and $L_2$ losses in Table \ref{tab:ab}.
We observe that the proposed $L_{1,2}$ achieves better results than $L_1$ and $L_2$ losses.

Also, in the proposed LandmarkGAN framework, the expression label is provided in the two generators, i.e., facial landmark generator $G_l$ and facial expression generator $G_e$.
To evaluate the influence of the expression label, we test with three different variants in Table \ref{tab:ab}, i.e., the expression label $l_e$ is only provided to the facial landmark generator (i.e., $l_e {\rightarrow} G_l$), the expression label $l_e$ is only provided to the facial expression generator (i.e., $l_e {\rightarrow} G_e$), the expression label $l_e$ is provided to both generators (i.e., $l_e {\rightarrow} G_l  \, \& \,  l_e {\rightarrow} G_e$).
As we can see in Table~\ref{tab:ab}, 
both ``$l_e {\rightarrow} G_l$'' and ``$l_e {\rightarrow} G_l  \, \& \,  l_e {\rightarrow} G_e$'' achieve almost the same performance, which proves that the emotion information mainly lies in the generated landmarks in the first stage, further validating the effectiveness of the proposed method.
Moreover, compared with ``$l_e {\rightarrow} G_l  \, \& \,  l_e {\rightarrow} G_e$'', the method ``$l_e {\rightarrow} G_e$'' achieves much worse results. The reason for this is that  the expression label is not provided in the first stage, resulting in unsatisfactory landmarks generated in the first stage, which in turn leads to unsatisfactory images produced in the second stage.

Lastly, the landmark based features consists in ``soft-crops'' of the image around each landmark location. 
Therefore, we also provide an ablation study in Table \ref{tab:ab} to show that this method (i.e., ``w/ soft-crops'') achieves better results than simply giving image and landmarks to the network (i.e., ``w/o soft-crops'').

\begin{table}[!t]
	\centering
	\caption{Results of expression recognition accuracy ($\uparrow$) with different training and testing settings.} 
    \resizebox{1\linewidth}{!}{%
		\begin{tabular}{lcccc} \toprule
			Train/Test  & Real/Real & Real/Syn. & Real+Nor./Real  & Real+Syn./Real         \\ \midrule
			JAFFE           & 34.44\%   & 25.83\%    & 35.84\% & \textbf{38.89\%} \\ 
			3DFE            & 46.75\%   & 35.33\%     & 47.64\% & \textbf{49.46\%}   \\ 
			FERG            & 99.93\%   & 95.63\%    & 99.95\% & \textbf{99.97\%}   \\ 	 \bottomrule	
	\end{tabular}}
	\label{tab:data_aug}
\end{table}

\begin{figure*}[!t]
	\centering
	\includegraphics[width=1\linewidth]{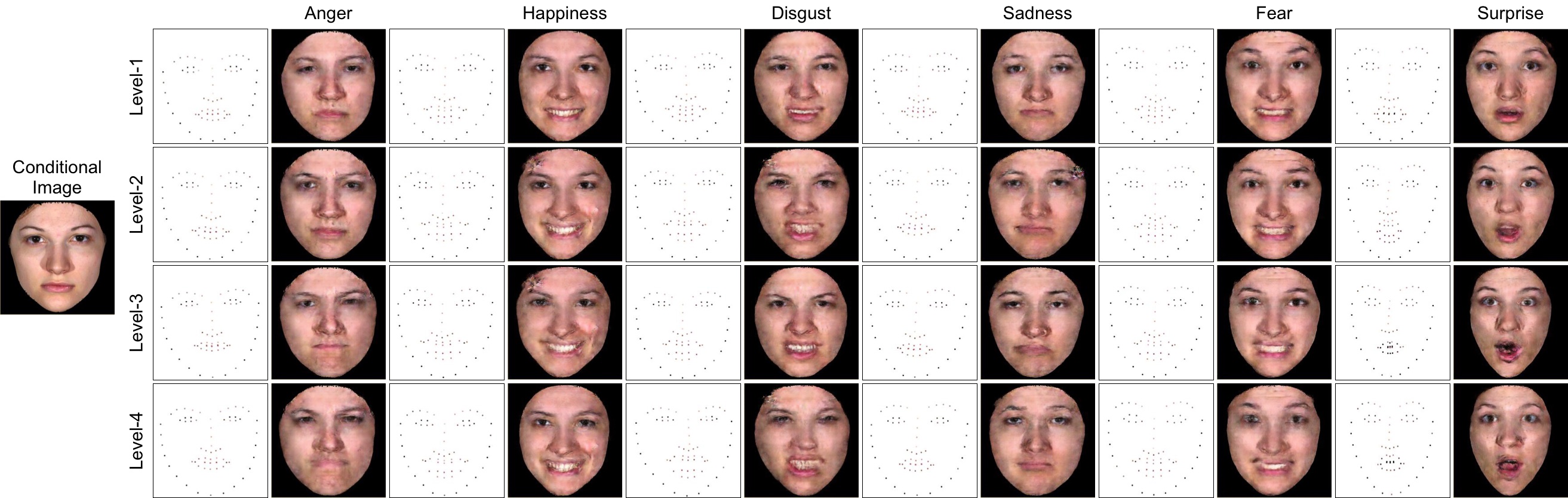}
	\caption{The generated landmark and expression with different intensity levels on 3DFE.}
	\label{fig:intensity_results}
\end{figure*} 

\begin{figure*}[!t]
	\centering
	\includegraphics[width=1\linewidth]{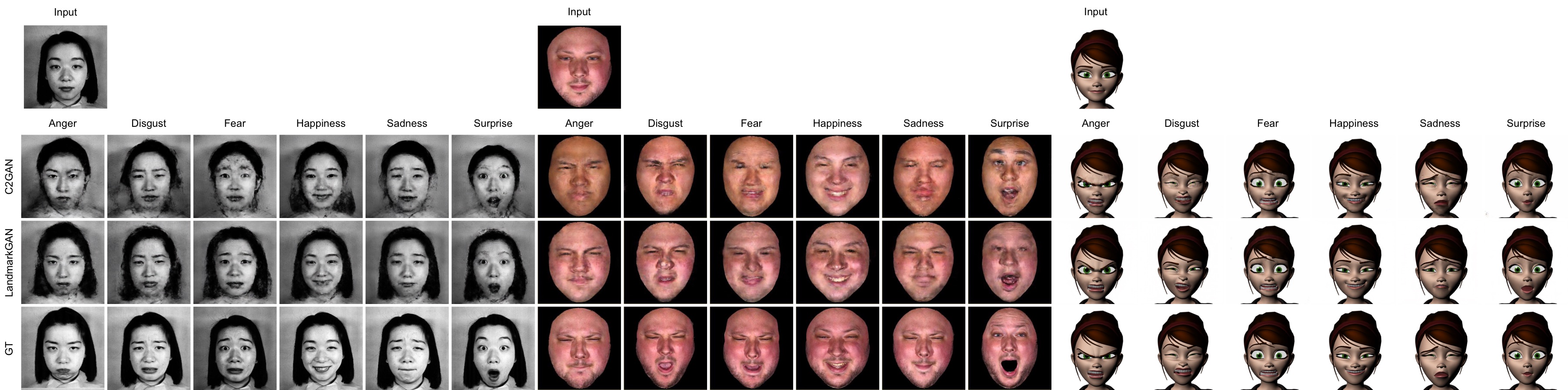}
	\caption{Different methods for facial expression-to-expression translation task on JAFFE, 3DFE, and FERG. From top to bottom: input, C2GAN \cite{tang2019cycle}, LandmarkGAN (Ours), and Ground Truth (GT). Note that the input of LandmarkGAN is only a single image. While C2GAN needs to input the target landmarks. The samples in this figure were randomly selected for visualization purposes.}
	\label{fig:comparision}
\end{figure*} 

\begin{figure*}[!t] 
	\centering
	\includegraphics[width=1\linewidth]{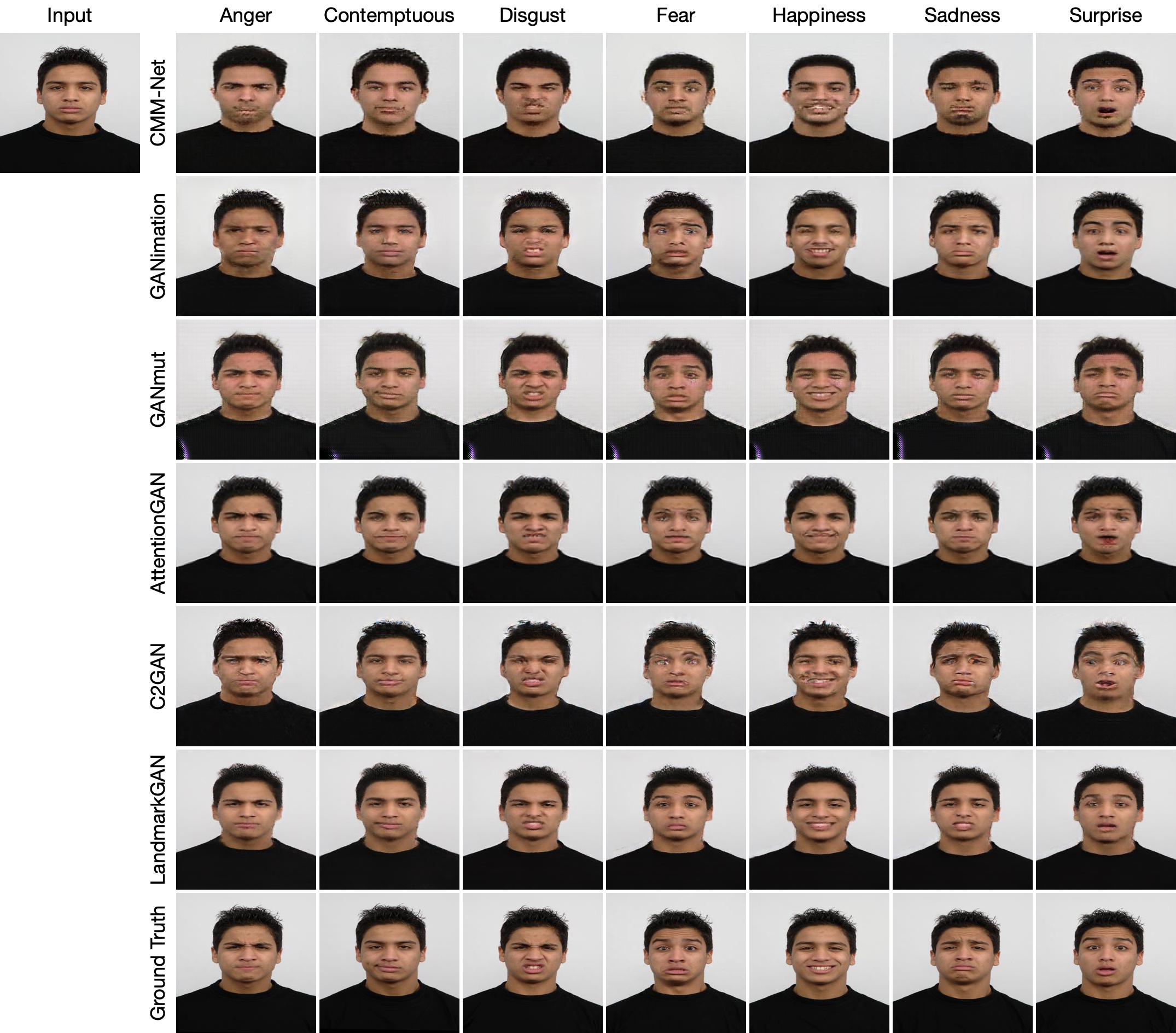}
	\caption{Different methods for facial expression-to-expression translation task on RaFD. From top to bottom:  CMM-Net \cite{wei2017every}, GANimation \cite{pumarola2018ganimation}, GANmut \cite{d2021ganmut}, AttentionGAN \cite{tang2021attentiongan}, C2GAN \cite{tang2019cycle},  LandmarkGAN (Ours), and Ground Truth (GT). The samples in this figure were randomly selected for visualization purposes.}
	\label{fig:rafd_comp}
\end{figure*} 

\begin{table*}[!t]
	\centering
	\caption{Quantitative results of different methods on JAFFE, 3DFE and FERG.}
	\begin{tabular}{lccccccccc} \toprule
		\multirow{2}{*}{Method}                           &\multicolumn{3}{c}{PSNR $\uparrow$}                          & \multicolumn{3}{c}{SSIM $\uparrow$}      & \multicolumn{3}{c}{IS $\uparrow$} \\ \cmidrule(lr){2-4} \cmidrule(lr){5-7} \cmidrule(lr){8-10}
		&JAFFE       & 3DEF   & FERG       &JAFFE   &3DFE    &FERG    &JAFFE     & 3DFE  & FERG \\ \midrule
	    GANimation~\cite{pumarola2018ganimation} &  15.3878 & 20.1987 & 37.9322 & 0.5176 & 0.8956 & 0.9875 & 1.3287 & 1.6720 & 1.4389 \\
	    {GANmut \cite{d2021ganmut}} & {15.6543} & {20.3445} & {38.1489} & {0.5221} & {0.8998} & {0.9887}& {1.3351} & {1.6806} & {1.4670} \\
		{AttentionGAN \cite{tang2021attentiongan}} & {16.1260} & {21.0475}& {38.3368}& {0.5267}& {0.9006}& {0.9896}& {1.3514}& {1.7169}& {1.5268}  \\
		C2GAN \cite{tang2019cycle}   &17.2442   &22.4289 & 38.4789  &0.5317  & 0.9147 & 0.9911 &1.5695     & 1.7531 &  1.5882 \\ 
		LandmarkGAN (Ours)                       &\textbf{17.2911}   &\textbf{23.8754 } & \textbf{38.9132} &\textbf{0.5392} & \textbf{0.9332} & \textbf{0.9932}  & \textbf{1.5708}    & \textbf{1.7984} & \textbf{1.6103}  \\		\bottomrule	
	\end{tabular}
	\label{tab:result1}
\end{table*}  

\noindent \textbf{Feature Visualization.}
We use t-SNE \cite{maaten2008visualizing} to visualize the feature learned from a pre-trained VGG-19 model~\cite{simonyan2014very}, which trained on a subset of the ImageNet dataset.
The feature is extracted after ``fc7'' fully connected layer, which is a 4,096-dim vector.
The results on the JAFFE, 3DFE, and FERG datasets are shown in Figure~\ref{fig:plot_id}.
Note that most generated samples are well separated, which demonstrates our LandmarkGAN can preserve identity information when converting facial expressions.

Moreover, we conduct face verification experiments on the RaFD dataset. The results of CMM-Net \cite{wei2017every}, GANimation \cite{pumarola2018ganimation}, GANmut \cite{d2021ganmut}, AttentionGAN \cite{tang2021attentiongan}, C2GAN \cite{tang2019cycle}, and the proposed LandmarkGAN are 0.902, 0.944, 0.948, 0.952, 0.968, and 0.981, respectively. 
These results show that the proposed method can better preserve the identity during translation.

\noindent \textbf{Data Augmentation.}
To show the generated images are useful for improving the performance of the facial expression recognition task, we employ the generated images as extra data for data augmentation purpose.
The intuition is that if the generated images are realistic, 
(i) the classifiers trained on both the real images and the generated images will be able to boost the accuracy of the real images (in this situation, the generated images work as augmented data.) and 
(ii) the classifiers trained on real images will also be able to classify the synthesized image correctly.
The results are listed in Table~\ref{tab:data_aug}.
Note that ``Real+Nor./Real'' means that the images produced by normal augmentation methods like adding noises, random rotation and crop for a fair comparison.
In this way the total number of images in training set should be same with the setting of ``Real+Syn./Real''. 
We can see that the recognition performance is boosted by adding the generated images by our method on all datasets.

\noindent \textbf{Control of Expression Intensity.}
Figure~\ref{fig:intensity_results} shows the results with different intensity levels of expressions on the 3DFE dataset.
\hao{For example, in the last column of Figure \ref{fig:intensity_results}, we can judge the intensity levels of the generated images by the size of the mouth opening. }
We present four levels of expression intensity from weak to strong.
To generate expressions with different intensities, Eq.~\eqref{eqn:stage_1} needs to be updated to, 
\begin{equation}
\begin{aligned}
\mathcal{L}_{cGAN}(G_l, D_l) & {=}  \mathbb{E}_{l} \left[\log D_l(l) \right]  \\
& +  \mathbb{E}_{x, z_1} \left[\log (1 - D_l(G_l(x, z_1, l_e, l_i))) \right], 
\end{aligned}
\end{equation}
where $l_i$ is the one-hot label of expression intensity.
Moreover, $\widehat{x}$ in Eq.~\eqref{eqn:clandmarkgan} needs to be updated to $\widehat{x}{=}[x, G_l(x, z_1, l_e, l_i), l_e, l_i]$.
Results in Figure~\ref{fig:intensity_results} validate that the proposed LandmarkGAN discovers the expression intensity by manipulating the facial landmark.
Note that Figure~\ref{fig:intensity_results} shows a very challenging task. We only need to input a neutral face image to generate face images with different intensity levels and expressions at the same time. 
To the best of our knowledge, our method is currently the only one that can achieve this goal.
\hao{Finally, we note that the stronger the expression, the better the facial expression classification result we will achieve.}

\noindent \textbf{Comparison with State-of-the-Art Methods.}
We compare the proposed LandmarkGAN with the most related facial expression generation models, i.e., CMM-Net \cite{wei2017every}, C2GAN \cite{tang2019cycle}, GANimation~\cite{pumarola2018ganimation}, AttentionGAN \cite{tang2021attentiongan}, and GANmut \cite{d2021ganmut}.
Note that C2GAN needs input a image and the target landmarks for generating the desired expressions.
For GANimation, which adopts facial action units (AUs) to guide the image generation process, AUs are the local components of the face which show muscle deformations. 
AttentionGAN is an unsupervised method, which can identify the most discriminative foreground contents and minimize the change of the background.
GANmut can generate a gamut of emotions by using only the categorical emotion labels.
Specifically,
we re-implemented the GANimation, GANmut, AttentionGAN, C2GAN on all the three JAFFE, 3DFE, and FERG datasets with the same training/inference configuration in their papers for fair comparisons.


\begin{table}[!t]
	\centering
	\caption{Quantitative results of different methods on RaFD. }
	\begin{tabular}{lcccc} \toprule
		Method                                  &  PSNR $\uparrow$  &   SSIM  $\uparrow$ & LPIPS  $\downarrow$  \\ \midrule
		CMM-Net \cite{wei2017every} & 18.6521 &0.8367 & 0.1282 \\
		GANimation \cite{pumarola2018ganimation}  & 20.8765 & 0.8432 & 0.1042 \\
		{GANmut \cite{d2021ganmut}} & {21.1348} & {0.8468} & {0.0982}\\
		{AttentionGAN \cite{tang2021attentiongan}} & {21.4586} & {0.8565} & {0.0967} \\
		C2GAN \cite{tang2019cycle}  & 21.9192 &  0.8618 & 0.0934 \\
		LandmarkGAN (Ours)            & \textbf{22.2132} & \textbf{0.8761}  & \textbf{0.0928} \\		 \bottomrule	
	\end{tabular}
	\label{tab:rafd}
\end{table}

\begin{figure*}[!ht]
	\centering
	\includegraphics[width=1\linewidth]{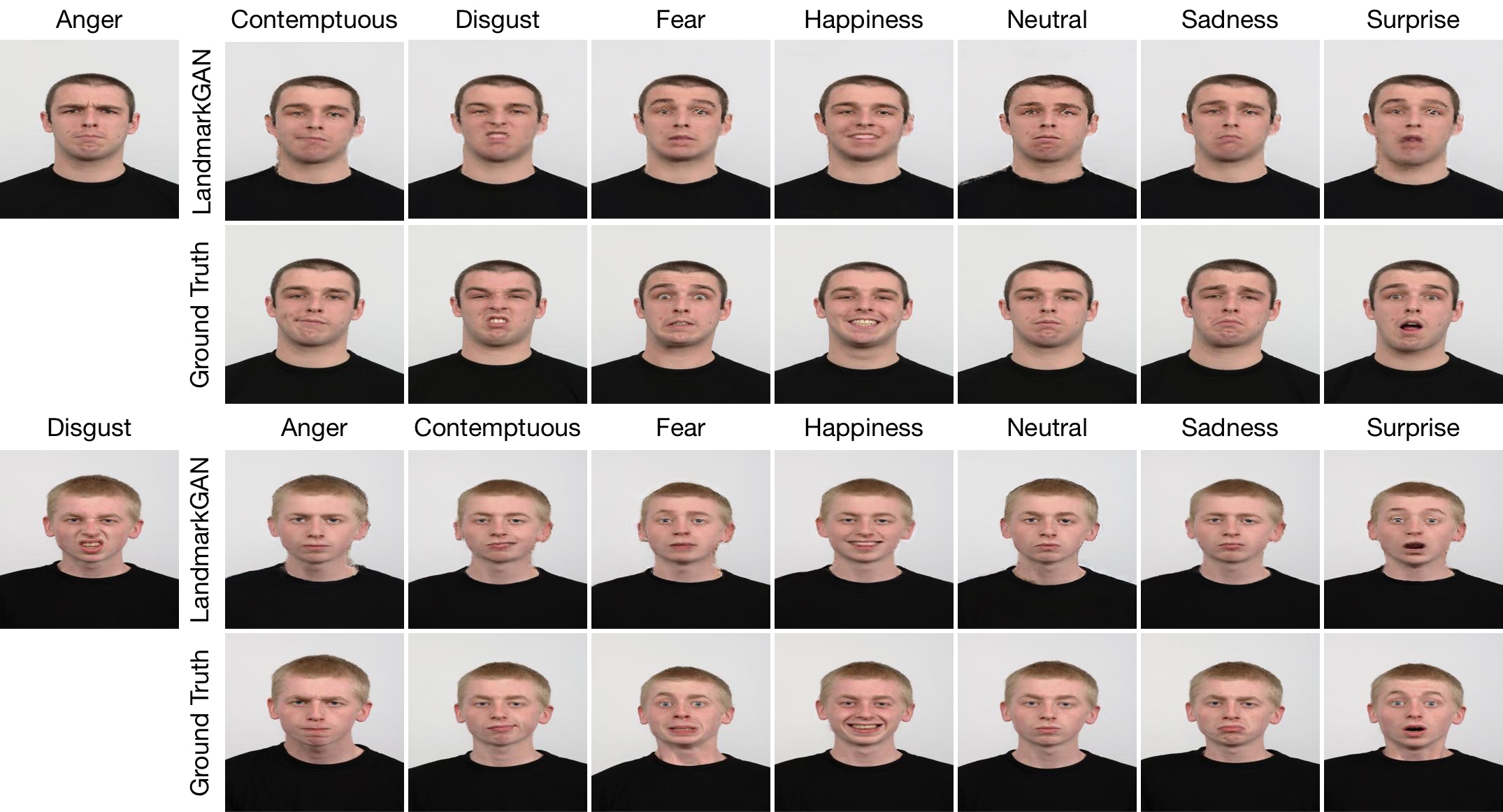}
	\caption{More results of the proposed LandmarkGAN on RaFD. In LandmarkGAN, we only need to input an arbitrary expression face and thus LandmarkGAN will generate other expressions.}
	\label{fig:rebuttal_result12}
\end{figure*} 

\begin{figure*}[!t]
	\centering
	\includegraphics[width=1\linewidth]{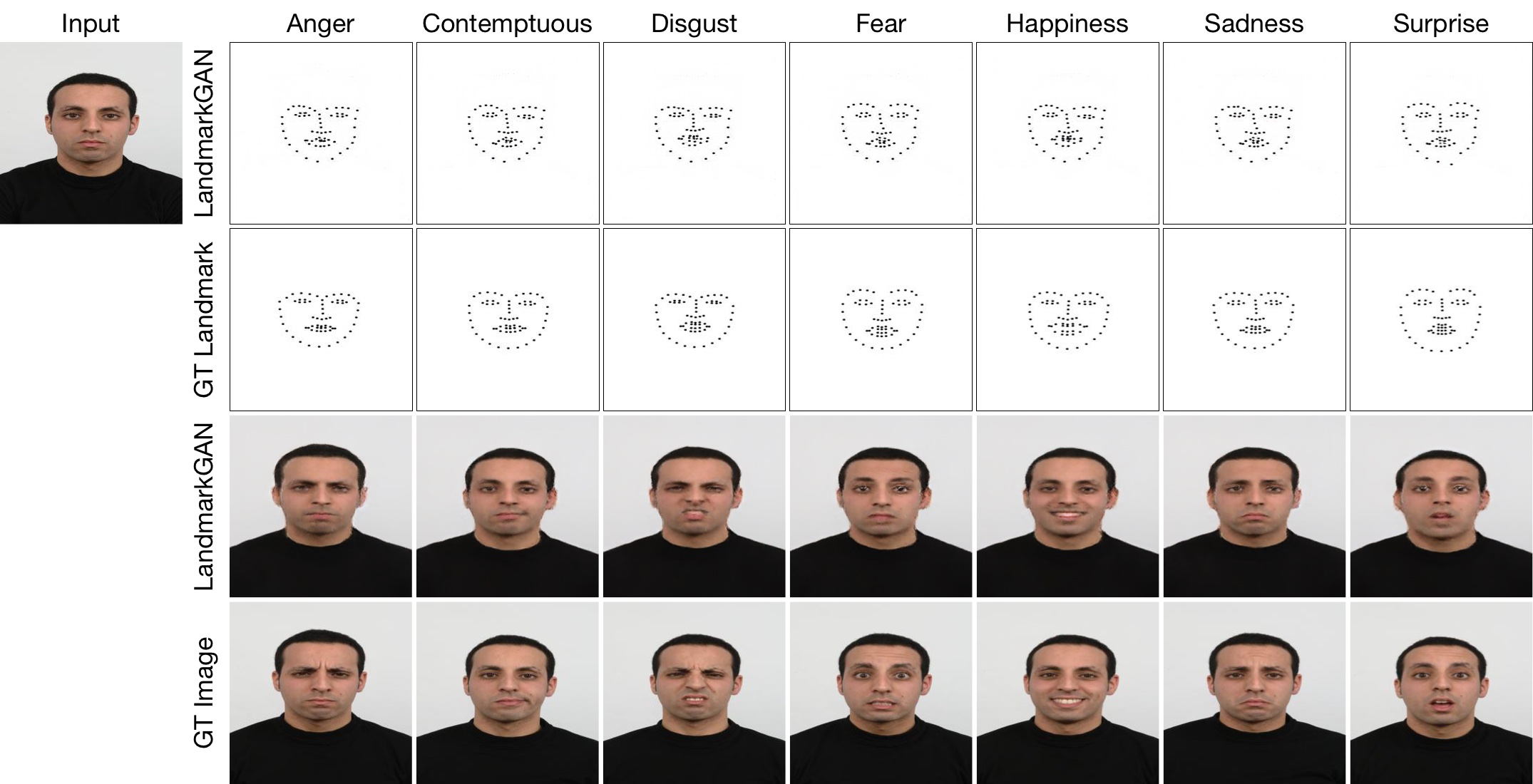}
	\caption{The generated landmarks and images of LandmarkGAN on RaFD. We only need to input a single image to obtain the diversity landmarks and expressions, while C2GAN needs to input a image and the GT landmarks to generate diversity expressions.}
	\label{fig:landmark}
\end{figure*} 

As we can see from Table~\ref{tab:result1} that the proposed LandmarkGAN consistently achieves better performance than C2GAN, GANimation, GANmut, and AttentionGAN on all the JAFFE, 3DEF, and FERG datasets.
We also show the visualization results compared to the state-of-the-art method, i.e., C2GAN, in 
Figure~\ref{fig:comparision}. We see that the proposed LandmarkGAN generate much better results than C2GAN on all the three datasets, which validates the effectiveness of our method. Moreover, the results of the RaFD dataset are shown in Figure~\ref{fig:rafd_comp} and Table~\ref{tab:rafd}.
We can see that the proposed LandmarkGAN achieves better results compared with the leading baselines, which validates the effectiveness of the proposed method.
Finally, in Figure~\ref{fig:rebuttal_result12} we show that the proposed LandmarkGAN can generate other different expressions by inputting only a anger or a disgust face.
\hao{The advantage of our proposed LandmarkGAN over these existing frameworks is that our framework is two-stage, the landmark generated in the first stage will help the second stage to generate a better image, and conversely, the image generated in the second stage will in turn help the first stage to generate better landmark. These two stages are trained in an end-to-end fashion so that we can  improve the generation results of the target landmark and image jointly.}

\begin{table}[!t]
	\centering
	\caption{AMT score ($\uparrow$) of different methods.}
	\begin{tabular}{lcccc} \toprule
		Method                                  &  JAFFE  &    3DFE  & FERG  &  RaFD  \\ \midrule
		C2GAN \cite{tang2019cycle}   & 1.2         &   13.3    & 70.9   & 34.2 \\
		LandmarkGAN (Ours)             & \textbf{2.5} &   \textbf{19.7}   & \textbf{82.9}    & \textbf{39.1} \\	 \bottomrule		
	\end{tabular}
	\label{tab:amt}
\end{table}  

\begin{table}[!t]
	\centering
	\caption{$L_2$ distance ($\downarrow$) for the landmark generation task.}
	\begin{tabular}{lcccc} \toprule
		Method                                  &  JAFFE     &    3DFE  & FERG  &  RaFD  \\ \midrule
		Sun et al. \cite{sun2017natural}             & 12.4         &    9.8     & 4.2     &  6.7 \\ 
		LandmarkGAN  (Ours)              & \textbf{11.2} &   \textbf{7.2}   & \textbf{3.5}    & \textbf{6.2} \\	 \bottomrule		
	\end{tabular}
	\label{tab:landmark}
\end{table}  

\noindent \textbf{User Study.} We follow the same setting in \cite{zhu2017unpaired,isola2017image} and conduct an perceptual studies to evaluate the quality of the generated images. 
Results are shown in Table \ref{tab:amt}.
We can see that the proposed LandmarkGAN consistently achieves better results compared with the leading method (i.e., C2GAN), which further validates the images generated by our method are more clear and visually plausible.

The metrics such as PSNR and LPIPS calculate the semantic distance between the real image and the generated image. The semantic distance between the images is not very large considering they are all faces. On the contrary, the AMT score in Table \ref{tab:amt} measures the generation quality from a perceptual level. The difference on the perceptual level is more obvious than on the semantic level, i.e., the generated images with small artifacts show a minor difference on the semantic level, while are being judged with a significant difference from the real images by humans.

\noindent \textbf{Visualization of Landmark Generation.}
We also compare landmark generation with \cite{sun2017natural}.
Similar to \cite{sun2017natural}, we employ $L2$ distance between detected and generated landmarks as the evaluation metric.
Quantitative results are provided in Table~\ref{tab:landmark}.
Note that the generated landmarks have a lower $L2$ distance than \cite{sun2017natural} as our multi-task learning strategy.
In Figure~\ref{fig:landmark}, we show some samples of the generated landmarks.
We see that the generated landmarks are very close to the target ones, which means LandmarkGAN can generate correct facial landmarks.
Overall, the proposed Landmark can generate facial landmarks and expressions simultaneously, which is not embraced by any existing work.

\begin{figure}[!t] 
	\centering
	\includegraphics[width=1\linewidth]{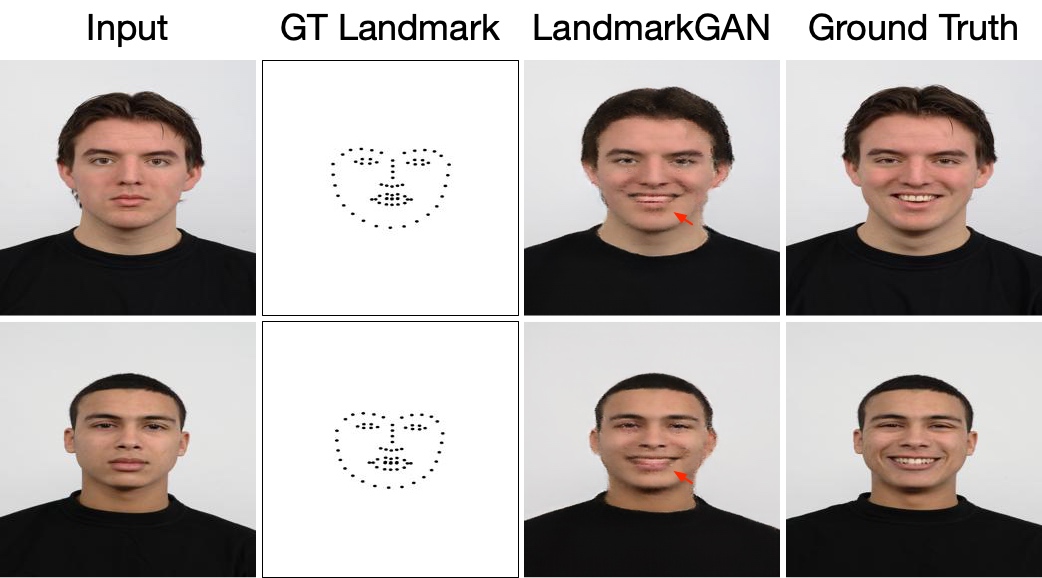}
	\caption{Failure cases on RaFD.}
	\label{fig:Failure}
\end{figure}

\noindent \textbf{\hao{Analysis of Failure Results}.}
\hao{As can be seen from Figure~\ref{fig:Failure}, our model cannot generate clear teeth, because we do not introduce any guidance in our framework to guide the generation of teeth. In future work, we will try to introduce the structural information of the teeth to generate realistic teeth. This could be very helpful for many generative tasks, such as image animation \cite{siarohin2019first} and audio-to-image translation \cite{prajwal2020lip}.}

\begin{figure}[!t] 
	\centering
	\includegraphics[width=1\linewidth]{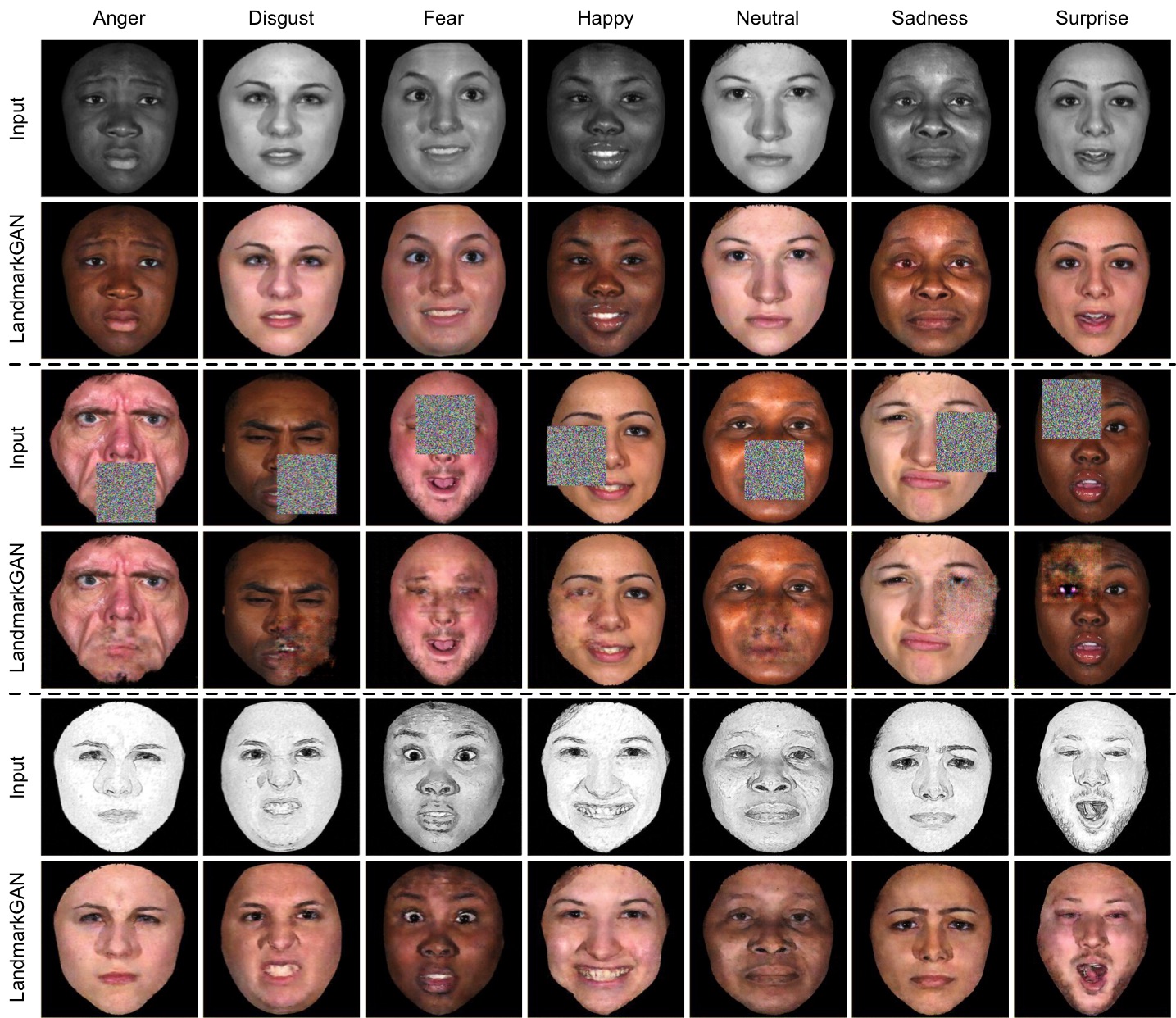}
	\caption{Results of LandmarkGAN on three generative tasks, i.e., face colorization (\textit{top}), face inpainting (\textit{middle}), and sketch-to-face translation (\textit{bottom}) on 3DFE.}
	\label{fig:applications}
\end{figure}

\noindent \textbf{Applications.}
The proposed LandmarkGAN is a task-irrelevant framework, thus we also show the results on other three generative tasks, i.e., face colorization, face inpainting, and sketch-to-face translation.
For face inpainting, images on 3DFE are re-scaled to $256 {\times} 256 {\times} 3$ first, then we set the mask size as $96 {\times} 96$ randomly to guarantee that at least one importance part of the face is missing.
For sketch-to-face translation, we use a public software to convert images from the 3DFE dataset into sketches, then to learn the mapping between sketch and image.
Results of the three tasks are shown in Figure~\ref{fig:applications}.
We see that the proposed LandmarkGAN generates reasonable results on the three different generative tasks, validating the generalization ability of the proposed LandmarkGAN.

\section{Conclusions}
\label{sec:conclusions}
We propose a novel LandmarkGAN for facial expression translation. 
The proposed LandmarkGAN can generate faces with arbitrary expressions and expression intensities.
The training of LandmarkGAN is comprised of two stages, (i) category-guided facial landmark generation and (ii) landmark-guided facial expression-to-expression translation.
Moreover, two novel losses include the landmark GAN loss, and the $L_{1,2}$ loss are proposed to learn the expression-to-expression mapping.
Experimental results demonstrate that the proposed LandmarkGAN is capable of generating higher quality faces with correct expression than the state-of-the-art approaches.
Lastly, since the proposed LandmarkGAN is task agnostic, it can be employed to other generative tasks such as expression recognition, face colorization, face inpainting, and sketch-to-face translation.

In this paper, we aim to generate basic emotions from a single image. Extending the proposed method to generate micro-expressions and dissimulated behavior is part of our further work.

\bibliographystyle{IEEEtran}
\bibliography{ref}

\begin{IEEEbiography}[{\includegraphics[width=1in,height=1.25in,clip,keepaspectratio]{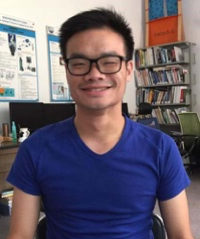}}]{Hao Tang}
is currently a Postdoctoral with Computer Vision Lab, ETH Zurich, Switzerland. 
He received the master’s degree from the School of Electronics and Computer Engineering, Peking University, China and the Ph.D. degree from Multimedia and Human Understanding Group, University of Trento, Italy.
He was a visiting scholar in the Department of Engineering Science at the University of Oxford. His research interests are deep learning, machine learning, and their applications to computer vision.
\end{IEEEbiography}

\begin{IEEEbiography}[{\includegraphics[width=1in,height=1.25in,clip,keepaspectratio]{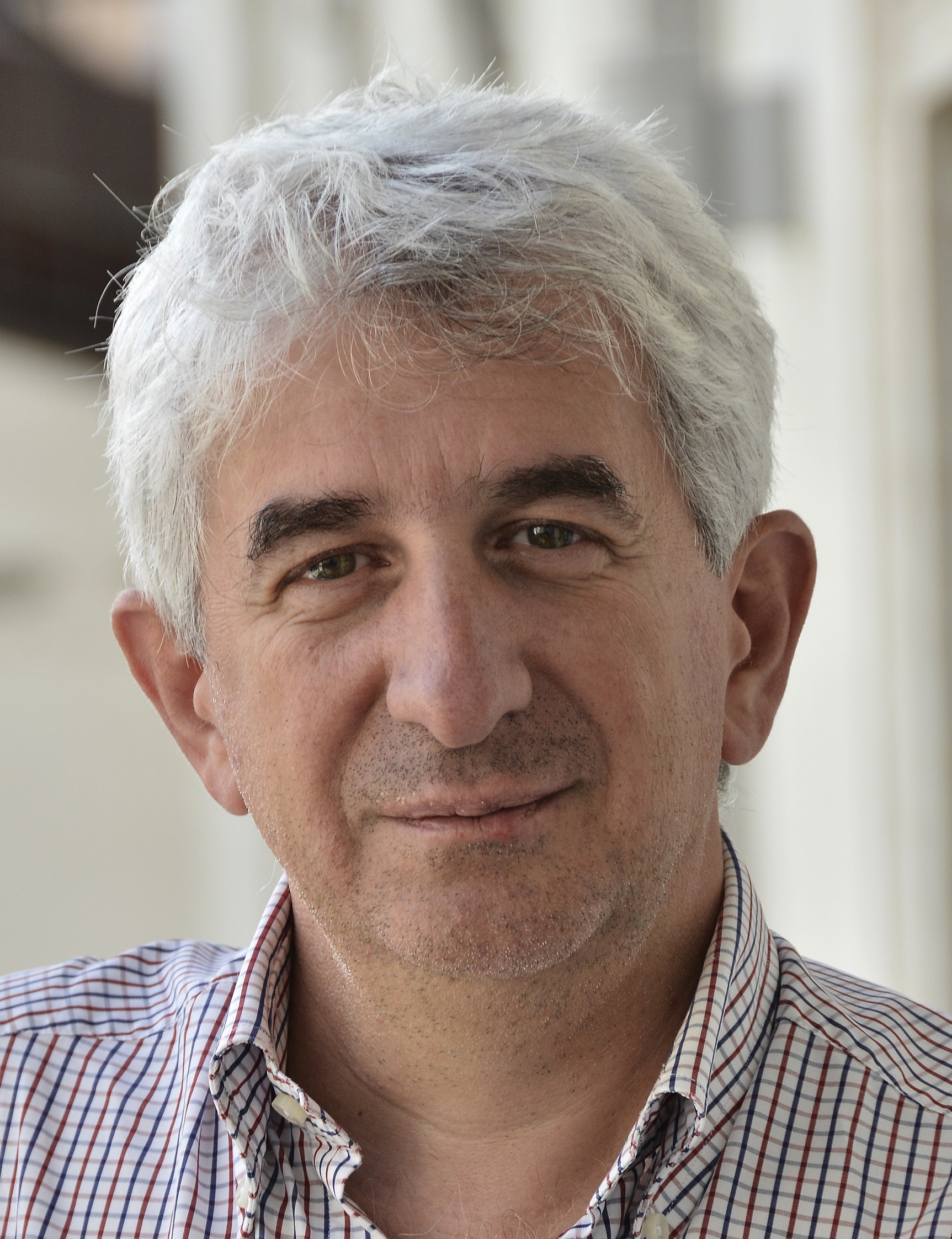}}]{Nicu Sebe} 
is Professor in the University of Trento, Italy, where he is leading the research in the areas of multimedia analysis and human behavior understanding. He was the General Co-Chair of the IEEE FG 2008 and ACM Multimedia 2013.  He was a program chair of ACM Multimedia 2011 and 2007, ECCV 2016, ICCV 2017 and ICPR 2020.  He is a general chair of ACM Multimedia 2022 and a program chair of ECCV 2024. He is a fellow of IAPR.
\end{IEEEbiography}

%
%
%
%
%




\end{document}